\documentclass{article} 
\usepackage{iclr2025_conference,times}

\usepackage{amsmath,amsfonts,bm}

\def\eqref#1{equation~\ref{#1}}

\def\1{\bm{1}}

\DeclareMathAlphabet{\mathsfit}{\encodingdefault}{\sfdefault}{m}{sl}
\SetMathAlphabet{\mathsfit}{bold}{\encodingdefault}{\sfdefault}{bx}{n}

\usepackage{verbatim}
\usepackage{enumitem}
\usepackage{makecell}
\usepackage[table]{xcolor}
\usepackage{amssymb}
\usepackage{multirow}
\usepackage{caption}
\usepackage{graphicx}
\usepackage{hyperref}
\usepackage{url}
\usepackage{amsmath}   
\usepackage{booktabs}  
\usepackage{tabularx}   
\newcolumntype{L}{>{\raggedright\arraybackslash}X}  

\definecolor{goodblue}{HTML}{0071bc}
\hypersetup{colorlinks=true,breaklinks,linkcolor=red,citecolor=goodblue}

\title{CAPEval: A Decoupled Caption Evaluation across Understanding and Generation}

\author{
Zhipeng Liu$^{1}$\thanks{Equal contribution. $\dag$ Corresponding authors.}
\quad Haochen Wang$^{1,2*\dag}$
\quad Zhaoxiang Zhang$^{1,2\dag}$
\\[2mm]
$^1$University of Chinese Academy of Sciences \\
$^2$Institute of Automation, Chinese Academy of Sciences \\[1pt]
{\small\texttt{
liuzp@ihep.ac.cn
}} \quad {\small\texttt{
\{wanghaochen2022, zhaoxiang.zhang\}@ia.ac.cn
}}\\[2mm]
\centerline{Project Page: \small{\url{https://liuzhipenggg.github.io/CAPEval}}}
}

\iclrfinalcopy 
\begin{document}

\maketitle

\begin{abstract}
Captions serve as a primary supervision signal for both multimodal understanding and text-to-image generation.
However, previous evaluations treat the caption quality as a single scalar objective, which conflates two distinct properties: (1) how much visual information a caption covers and (2) how reliably the image supports its stated claims. 
To this end, we design a \textit{decoupled} caption evaluation benchmark, \textbf{CAPEval} (\textbf{C}overage \textbf{A}nd \textbf{P}recision \textbf{Eval}uation), with human-written ground-truth captions and human-verified atomic checklist items. 
Specifically, CAPEval decomposes caption quality into \emph{Coverage} $C$ and \emph{Precision} $P$. 
The former quantifies how thoroughly a caption covers ground-truth factual content, while the latter reflects the factual correctness rate of all claims expressed in the caption.
We select 10 captioners and further conduct controlled downstream \textit{end-to-end} experiments with them from four model families, where the caption source is the only variable.
Empirically, we find a consistent \textit{task-dependent} dissociation: Coverage serves as the stronger correlate for understanding performance, whereas Precision acts as the dominant predictor for generation performance.
This decoupled evaluation paradigm not only delivers a more fine-grained diagnosis of caption quality, but also offers actionable guidance for selecting and optimizing captioners tailored to different downstream tasks.
%
%
\end{abstract}

\section{Introduction}

\begin{figure}[h]
    \centering
    \includegraphics[width=\linewidth]{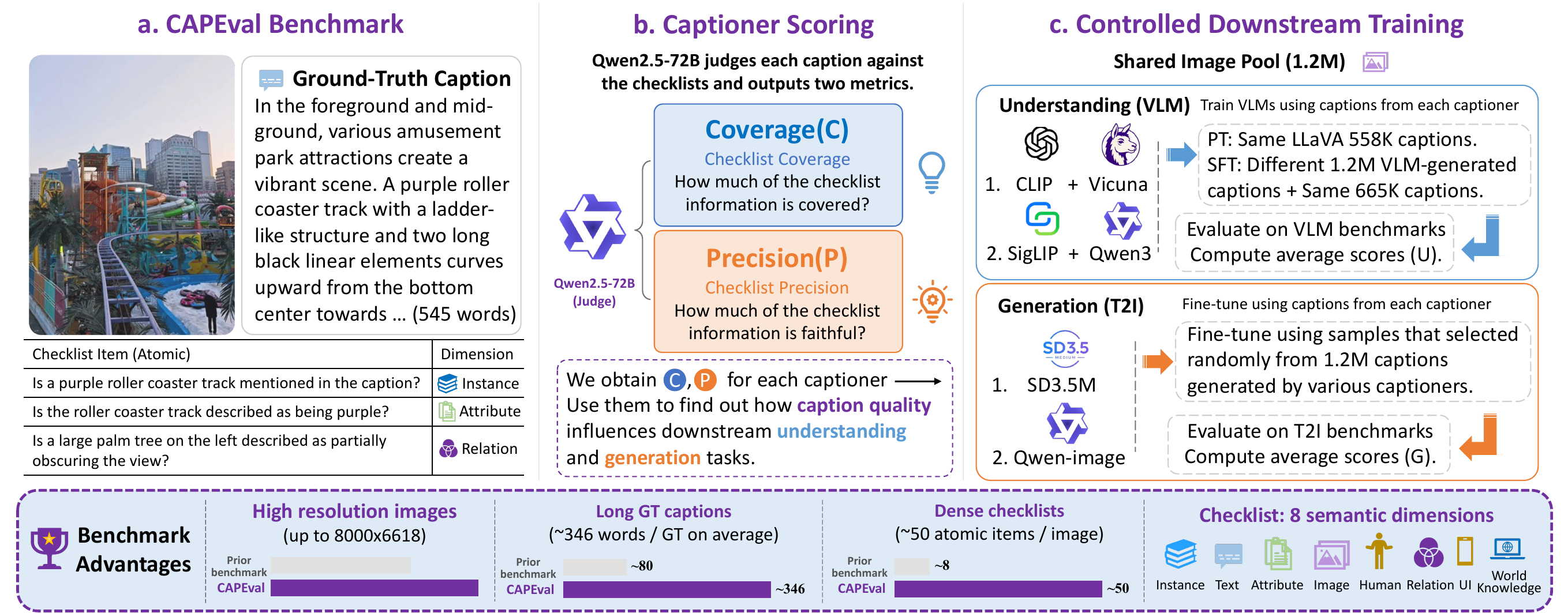}
    \vspace{-15pt}
    \caption{Overview of CAPEval.
    \textbf{(a)} CAPEval constructs fine-grained semantic checklists from high-resolution (up to 8K) images, long-form ground-truth captions, and dense atomic checklist items spanning eight semantic dimensions.
    \textbf{(b)} Each candidate caption is evaluated against the atomic checklist items by a judge model to obtain \textit{Coverage} ($C$) and
    \textit{Precision} ($P$).
    \textbf{(c)} Captions produced by different captioners are then used in controlled VLM training and T2I fine-tuning experiments, where the caption source is the only varying factor, yielding aggregate understanding score $U$ and generation score $G$ for each captioner.
    }\label{fig:introduction_1}
    \vspace{-10pt}
\end{figure}

To achieve comprehensive understanding and precise text-instruction following capabilities, large-scale image caption corpora~\citep{schuhmann2022laion, gadre2023datacomp, zhang2025low, dong2025scalable, wei2025hq, farina2026datacomp, li2026claimdiff} are widely used for training \textit{both} vision-language models (VLMs)~\citep{radford2021learning, jia2021scaling, alayrac2022flamingo, li2023blip, wang2025ross3d, wang2025reconstructive, lei2025scalability, wang2026traceable, wang2025grasp, wang2025vgr, liu2026motionatlas, zhang2026actor} and text-to-image (T2I) generators~\citep{rombach2022high, wu2025qwen, team2025longcat}.
Under such a setting, textual descriptions, \textit{i.e.}, captions, serve as the primary supervision signal for aligning visual content with language.
Recent studies reveal that caption quality has become an explicit data variable that can substantially affect downstream multimodal capabilities~\citep{chen2024sharegpt4v, zhang2025low, yang2026caprl++}.

However, how caption-quality properties affect multimodal learning remains insufficiently explored.
More specifically, \textit{whether downstream multimodal models benefit more from captions that cover a broader range of visual content (i.e., coverage), or from captions that make fewer but more reliable claims (i.e., precision), and whether the conclusion differs between understanding and generation?}
To systematically evaluate this, we design \textbf{CAPEval} (\textbf{C}overage \textbf{A}nd \textbf{P}recision \textbf{Eval}uation), based on the following two principles:

\begin{enumerate}
\item \textbf{Decoupled Evaluation of Coverage and Precision.} 
%
%
Both conventional single-scalar captioning metrics~\citep{papineni2002bleu, vedantam2015cider, hessel2021clipscore} and recent enhanced evaluation methods~\citep{gao2026gavel, liu2026unison} conflate coverage and reliability of descriptions into one unified score without disentanglement. 
We, on the contrary, aim to explicitly \textit{decompose} caption quality into Coverage and Precision to systematically explore how the relative importance may \textit{differ} across downstream scenarios.
\item \textbf{End-to-End Captioner Evaluation.}
Prior benchmarks~\citep{onoe2024docci, dong2024benchmarking} evaluate captions as standalone outputs, leaving unclear how caption quality propagates into downstream tasks.
We, instead, adopt a fully \textit{end-to-end} protocol: \textit{each captioner directly generates training captions}, with the caption source as the only variable across identical pipelines.
This design links intrinsic caption quality (Coverage and Precision) to downstream VLM and T2I performance, converting captioner evaluation into a direct measure of training data utility.

\end{enumerate}

\begin{figure}[t]
    \centering
    \includegraphics[width=\linewidth]{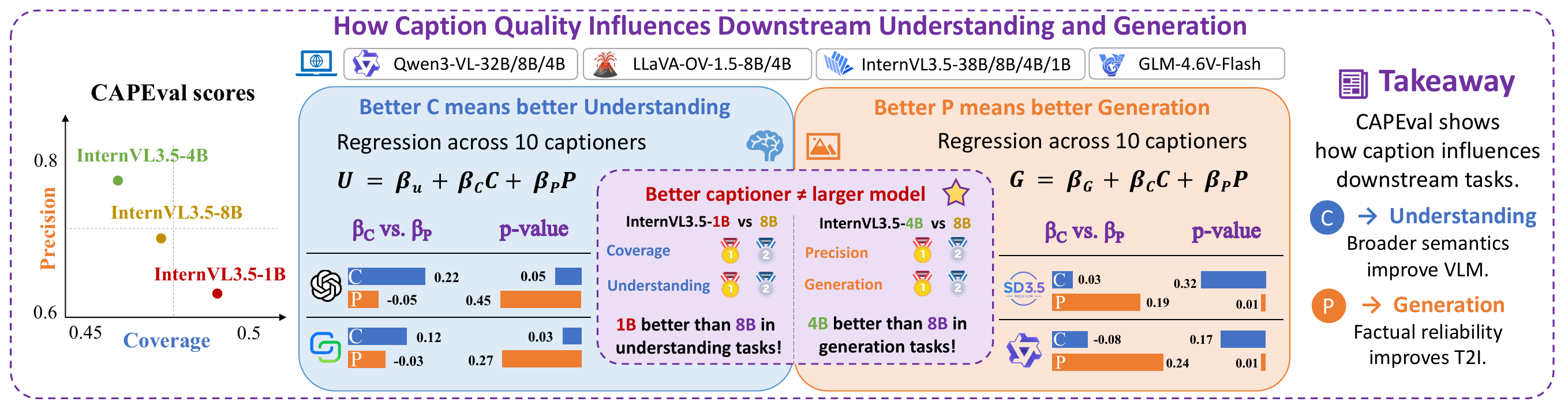}
    \vspace{-20pt}
    \caption{Relating CAPEval scores to downstream performance reveals a task-dependent asymmetry: \textit{broader semantic Coverage drives VLM understanding}, whereas \textit{higher factual Precision drives T2I generation}.
    This pattern is confirmed by both the InternVL3.5 case study and regression analysis across ten captioners in Section~\ref{sec:downstream}.
    }
    \label{fig:introduction_2}
    \vspace{-10pt}
\end{figure}

Concretely, CAPEval operates as a four-stage pipeline: (1) dense atomic checklists are constructed from selected images and long-form human-annotated ground-truth (GT) captions as shown in Figure~\ref{fig:introduction_1}\textcolor{red}{a}; (2) per-captioner Coverage and Precision scores are computed by a judge model as shown in Figure~\ref{fig:introduction_1}\textcolor{red}{b}; (3) captions from each captioner serve as the sole varying supervision signal in controlled VLM pretraining and T2I fine-tuning, yielding an aggregate understanding score $U$ and generation score $G$ per captioner as shown in Figure~\ref{fig:introduction_1}\textcolor{red}{c}; and (4) CAPEval fits ordinary least squares (OLS) regressions $U\ \text{or}\ G = \beta_0 + \beta_C C + \beta_P P$ across captioners, and systematically examines the \textit{significance} ($p$-values) and \textit{magnitude} ($\beta_C$ and $\beta_P$) to characterize how Coverage and Precision contribute to downstream performance, as shown in Figure~\ref{fig:introduction_2}.
Our experiments reveal three key findings:

\begin{enumerate}
    \item \textbf{Caption quality profile matters more than model scale.}
    Smaller captioners can produce more useful supervision when their caption properties better align with the downstream objective.
    For instance, within the InternVL3.5 family, the smallest model (1B) achieves a higher understanding score than the 8B model (avg.\ $U = 58.5$ vs.\ $57.3$) and also exhibits higher Coverage ($C = 48.3$ vs.\ $46.5$).
    Similarly, the 4B model outperforms its larger 8B counterpart on generation (avg.\ $G = 71.6$ vs.\ $71.0$) while attaining higher Precision ($P = 73.5$ vs.\ $72.6$).
    These within-family observations suggest that understanding performance may be more closely associated with Coverage, whereas generation performance may be more closely associated with Precision.
    We next examine whether these relationships hold more broadly across captioners through systematic regression analysis.

    \item \textbf{Understanding is dominated by Coverage.} For understanding, Coverage is the stronger and consistent predictor ($\hat{\beta}_C = +0.118$, $p$-value $= 0.026$ on SigLIP-Qwen3; $\hat{\beta}_C = +0.215$, $p$-value $= 0.047$ on CLIP-Vicuna), while Precision is not in both pipeline ($p$-value $\geq 0.274$ in either pipelines), indicating that broad semantic coverage is the dominant factor. We further find that unlike general understanding, hallucination performance is dominated by Precision ($p$-value $= 0.124$), while Coverage is less significant ($p$-value $= 0.914$).

    \item \textbf{Generation is driven by Precision.} Across ten captioners, Precision is the only statistically significant predictor of generation quality ($\hat{\beta}_P = +0.189$, $p$-value $< 0.001$ on SD3.5M; $\hat{\beta}_P = +0.235$, $p$-value $< 0.001$ on Qwen-Image), while Coverage is not statistically significant in either pipeline (both $p$-values $\geq 0.171$), indicating that factual precision is the dominant factor for text-to-image generation.

\end{enumerate}

These results reveal a task-relative view of caption quality: the dimension that matters most is determined by the downstream objective rather than by captioner scale or aggregate quality score.
CAPEval provides a practical foundation for task-aware caption data curation: understanding models should be trained on captions with broad semantic coverage, while generation models benefit from captions with high factual precision.

\section{Related Work}

Captions serve as a foundational supervision signal across \textit{both} multimodal understanding and text-to-image generation, and much work has consistently demonstrated that improving caption quality drives measurable gains in downstream model performance.

In multimodal understanding, captions act as the core supervision signal for cross-modal alignment and capability building.
Qwen-VL~\citep{bai2023qwen} first establishes captions as a foundational training anchor, unifying image description, visual grounding, and text reading abilities by aligning image–caption–box tuples. 
ShareGPT4V~\citep{chen2024sharegpt4v} further verifies that caption quality directly determines downstream gains: replacing coarse alt-text with detailed, information-rich captions consistently boosts modality alignment, instruction following, and visual reasoning across model backbones. 
Most recently, the GLM series~\citep{hong2025glm} refines caption properties via factuality-centered recaptioning and data filtering, lifting caption density, image–text relevance, and factual accuracy to strengthen general multimodal reasoning.

In text-to-image generation, captions act as the primary conditioning interface that enables semantic control over visual outputs.
State-of-the-art generation systems have further identified caption quality as a critical performance lever: Stable Diffusion 3~\citep{esser2024scaling} mixes original captions with CogVLM-generated synthetic captions~\citep{wang2024cogvlm}.
DALL-E 3~\citep{betker2023improving} \textit{recaptions} training images with a dedicated caption model to enhance prompt following.
Qwen-Image~\citep{wu2025qwen} prioritizes high-quality image-text data curation to boost overall capability.

Despite this widespread recognition of caption quality, existing evaluation paradigms, from conventional reference-based lexical matching to recent fine-grained checklist- and rubric-based assessment, face two core limitations that block a fine-grained, actionable understanding of how caption properties impact downstream learning. 
Early benchmarks built on Flickr8k~\citep{hodosh2013framing}, Flickr30k~\citep{young2014image} and MS COCO Captions~\citep{chen2015microsoft} quantify quality via metrics like BLEU~\citep{papineni2002bleu}, METEOR~\citep{banerjee2005meteor} and CIDEr~\citep{vedantam2015cider}, which measure lexical overlap between generated captions and human-written ground-truth references. 
As long-form detailed captioning has become the mainstream, newer works shifted to factuality-aware, fine-grained diagnosis free from rigid reference constraints: DOCCI~\citep{onoe2024docci} and DetailCaps~\citep{dong2024benchmarking} adopt atomic checklists to evaluate dense visual coverage.
PerceptionRubrics~\citep{wei2026perceptionrubrics} uses structured scoring rubrics.
GAVEL~\citep{gao2026gavel} and Unison~\citep{liu2026unison} further extend the scope to grounded error localization and joint understanding-generation consistency.

While we have witnessed this methodological evolution, no existing benchmark disentangles caption quality into separate coverage and reliability dimensions.
All frameworks either conflate the two into a single scalar score \textit{without explicit decomposition}. 
Moreover, nearly all evaluation paradigms treat captions as standalone generation outputs, rather than assessing their actual utility as training supervision signals through \textit{end-to-end} downstream experiments. 
To fill these gaps, we present CAPEval, a decoupled caption evaluation benchmark that explicitly decomposes caption quality into Coverage and Precision, paired with a fully end-to-end evaluation protocol that directly measures the downstream utility of captions across both understanding and generation.

\section{CAPEval Design}

\begin{figure}[t]
    \centering
    \includegraphics[width=1.0\linewidth]{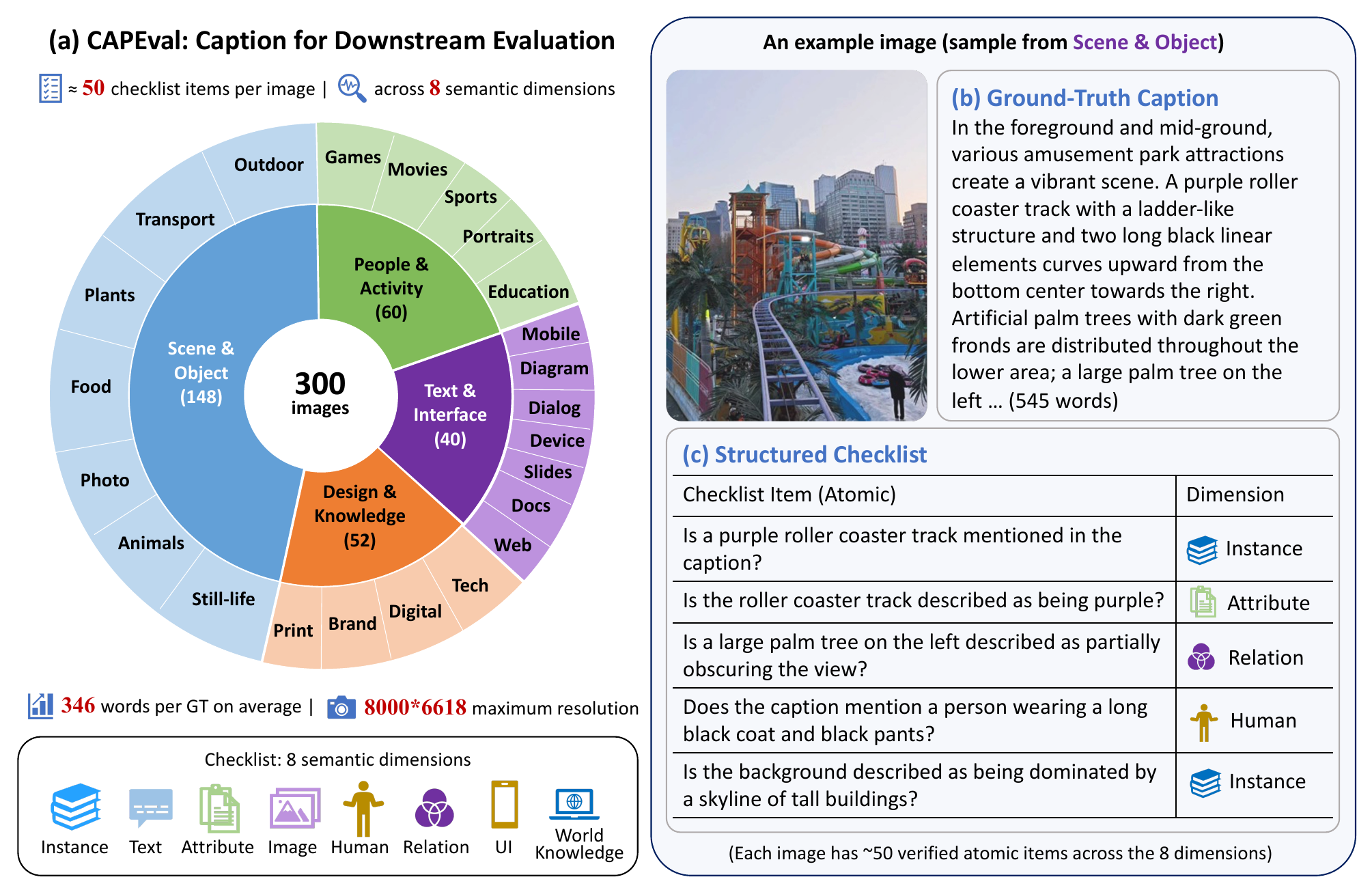}
    \vspace{-20pt}
    \caption{CAPEval image collection and annotation design.
    \textbf{(a)}~Hierarchical taxonomy and distribution of 300 images across four domains and subcategories.
    \textbf{(b)}~Example ground-truth caption (545 words) for a sample image.
    \textbf{(c)}~Corresponding structured semantic checklist, showing atomic items and their assigned dimensions. The complete information of GT caption and checklists for this example is provided in Appendix~\ref{sec:gt_checklist_example}.}
    \label{fig:capeval_design}
    \vspace{-10pt}
\end{figure}

\subsection{Overview}

We introduce \textbf{CAPEval} (\textbf{C}overage \textbf{A}nd \textbf{P}recision \textbf{Eval}uation), a unified framework for systematically connecting caption quality with downstream multimodal capability.
CAPEval is grounded in structured semantic checklists: each image is paired with a human-generated GT caption that is decomposed into atomic checklist items spanning eight dimensions (Figure~\ref{fig:introduction_1}\textcolor{red}{a}). Given a candidate caption, Qwen2.5-72B~\citep{hui2024qwen2} judges each checklist item as \textit{correct}, \textit{wrong}, or \textit{not mentioned}, based solely on the caption text without access to the image (Figure~\ref{fig:introduction_1}\textcolor{red}{b}). Based on these checklist annotations, CAPEval decomposes caption quality into two dimensions:

\begin{itemize}
    \item \textbf{Coverage $C$}: the fraction of GT checklist items that are mentioned by the candidate caption, measuring semantic completeness.
    \item \textbf{Precision $P$}: the fraction of mentioned checklist items that are factually correct, measuring factual reliability.
\end{itemize}

Rather than evaluating captions in isolation, CAPEval connects intrinsic caption quality with downstream performance through controlled training experiments. As illustrated in Figure~\ref{fig:introduction_1}\textcolor{red}{c}, captions generated by each captioner serve as the sole varying supervision signal across two VLM pretraining pipelines and two T2I fine-tuning pipelines.
By jointly analyzing Coverage, Precision, and downstream performance, CAPEval establishes an explicit bridge between caption evaluation and multimodal performance, enabling systematic identification of which caption quality dimension is most predictive for each downstream objective, as summarized in Figure~\ref{fig:introduction_2}.

\subsection{Benchmark Construction}

\begin{figure}[t]
    \centering
    \includegraphics[width=\linewidth]{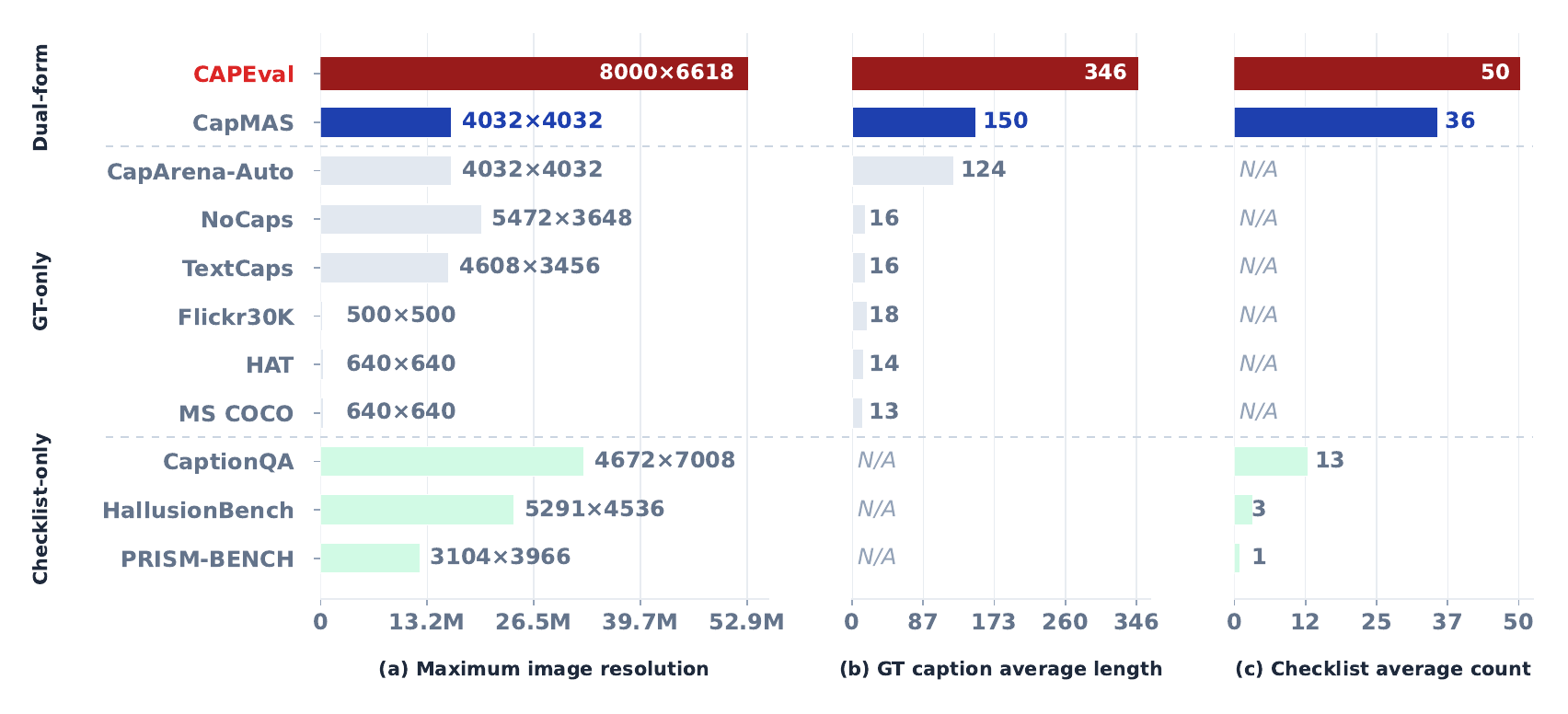}
    \vspace{-20pt}
    \caption{Comparison of CAPEval with existing benchmarks across \textbf{(a)}~maximum image resolution (Additional details are provided in Appendix~\ref{app:Image Resolution Distribution}), \textbf{(b)}~average GT caption length, and \textbf{(c)}~average checklist items per image (Additional details are provided in Appendix~\ref{app:Checklist dimension}). Benchmarks are grouped into Dual-form, GT-only, and Checklist-only. CAPEval leads on all three axes.}
    \label{fig:capeval_comparison}
    \vspace{-10pt}
\end{figure}

\textbf{Image Collection.}
We select 300 images covering four visually diverse domains, each further divided into several subcategories, as shown in Figure~\ref{fig:capeval_design}. Images are sourced from publicly available web sources and real-world photographs, spanning a wide range of resolutions and aspect ratios. As shown in Figure~\ref{fig:capeval_comparison}\textcolor{red}{a}, CAPEval's maximum image resolution exceeds that of prior benchmarks~\citep{
lee2024toward,
cheng2025caparena,
agrawal2019nocaps,
sidorov2020textcaps,
young2014image,
petryk2024aloha,
lin2014microsoft,
yang2025captionqa,
guan2023hallusionbench,
fang2025flux}.

\textbf{GT Captions.}
Each image is paired with a GT caption written by a human annotator following explicit guidelines: captions must cover all present elements across five dimensions (visual subjects, in-image text verbatim, aesthetic and photographic attributes, portrait attributes, and special image properties), describe only objectively visible content without speculation or repetition, and pass review by a second annotator before finalization. These guidelines produce captions that are substantially longer and more factually dense than those in prior benchmarks, as shown in Figure~\ref{fig:capeval_comparison}\textcolor{red}{b}.

\textbf{Semantic Checklists.}
Each GT caption is decomposed into a structured checklist of atomic statements spanning eight semantic dimensions. All items are manually verified for factual grounding and correct dimension assignment. As shown in Figure~\ref{fig:capeval_comparison}\textcolor{red}{c}, CAPEval provides very dense checklists, with a complete annotation protocol provided in Appendix~\ref{app:annotation}.

\subsection{CAPEval Metrics}
\label{app:CAPEval_metrics}

CAPEval decomposes caption quality into two axes: \emph{Coverage ($C$)}, measuring the fraction of GT facts the caption attempts to mention, and \emph{Precision ($P$)}, measuring the fraction of mentioned facts that are correct.
For each image, the caption generated by each captioner is compared against the GT checklist using Qwen2.5-72B as the judge.
Every checklist item is labeled as \emph{yes}, \emph{no}, or \emph{not mentioned}, indicating whether the item is correctly described, incorrectly described, or omitted by the caption.
Let $N_{\text{yes}}$, $N_{\text{no}}$ and $N_{\text{nm}}$ denote the numbers of \emph{yes}, \emph{no} and \emph{not mentioned} items.
Then Coverage is computed by
\begin{equation}
C =
\frac{100 \times \left(N_{\text{yes}} + N_{\text{no}}\right)}
{N_{\text{yes}} + N_{\text{no}} + N_{\text{nm}}}.
\end{equation}

Precision is computed by
\begin{equation}
P =
\frac{100 \times N_{\text{yes}}}
{N_{\text{yes}} + N_{\text{no}}}.
\end{equation}

\subsection{Controlled Training and Evaluation}

To study how caption quality affects downstream vision-language understanding and text-to-image generation, we conduct controlled training experiments in which the caption source is the only variable. We evaluate ten captioning models from four model families: InternVL3.5 with 1B, 4B, 8B, and 38B variants~\citep{wang2025internvl3}; Qwen3-VL with 4B, 8B, and 32B variants~\citep{bai2025qwen3}; LLaVA-OneVision-1.5 with 4B and 8B variants~\citep{an2025llava}; and GLM-4.6V-Flash~\citep{hong2025glm}. For each captioner, we construct a corresponding training corpus by replacing only the caption supervision. This design allows performance differences to be attributed primarily to differences in caption quality rather than confounding factors in data composition or training configuration.

\textbf{Vision-Language Understanding.}
We fine-tune two VLM pipelines to find out how caption quality influences vision-language understanding. The first pipeline uses CLIP-ViT-Large-Patch14-336~\citep{radford2021learning} as the visual encoder and Vicuna-7B-v1.5~\citep{touvron2023llama} as the language model. The second pipeline uses SigLIP-SO400M-Patch14-384~\citep{zhai2023sigmoid} as the visual encoder and Qwen3-4B~\citep{yang2025qwen3} as the language model.
For both pipelines, we follow the standard two-stage training procedure. In Stage 1, we train the visual-language projector on the same LLaVA 558K pretraining set~\citep{liu2023visual} with fixed captions, so this stage introduces no variation. In Stage 2, we vary only the caption supervision while following the ShareGPT4V data composition, sampling 1.2M images from COCO~\citep{lin2014microsoft}, SAM~\citep{kirillov2023segment}, and LLaVA/LCS~\citep{liu2023visual}. For each caption source, the generated captions are combined with the same 665K instruction-following mixture used in LLaVA-1.5~\citep{liu2024improved}.
Across all runs, we keep the model architecture, image pool, optimization settings, and training schedule fixed; only the caption source for the 1.2M image pool is changed. This controlled design isolates the downstream effect of caption-quality properties on VLM performance.

We evaluate the resulting VLMs on 16 benchmarks covering four capability dimensions: (1)~general multimodal understanding on MME~\citep{fu2026mme}, MMBench (EN and CN)~\citep{liu2024mmbench}, SEED-Bench~\citep{li2023seed}, and MMMU~\citep{yue2024mmmu}; (2)~multimodal reasoning on ScienceQA~\citep{lu2022learn}, AI2D~\citep{kembhavi2016diagram}, and RealWorldQA~\citep{ai2024grok}; (3)~visual perception on MMStar~\citep{chen2024we}, MMVP~\citep{tong2024eyes}, CV-Bench (2D and 3D)~\citep{tong2024cambrian}, and OCRBench~\citep{liu2024ocrbench}; and (4)~hallucination robustness on POPE~\citep{li2023evaluating}, HallusionBench~\citep{guan2023hallusionbench}, and AMBER~\citep{wang2023amber}.

\textbf{Text-to-Image Generation.}
We fine-tune two diffusion-based T2I models to test whether caption-quality effects also transfer to image generation. Specifically, we fine-tune Stable Diffusion 3.5 Medium (SD3.5M)~\citep{esser2024scaling} on 50K images and Qwen-Image~\citep{wu2025qwen} on 100K images. The two training subsets are sampled from the same 1.2M image pool and the captions are generated from each captioner.
Following the same controlled-variable principle, we construct one training set for each caption source by replacing only the caption supervision while keeping the sampled images unchanged. For each T2I backbone, all runs use the same model architecture, image subset, optimization settings, and training schedule. Thus, the only variable is the caption source which allows us to evaluate whether different caption-quality properties affect image generation.

We evaluate the fine-tuned T2I models on three benchmarks targeting complementary aspects of generation quality: (1)~compositional generation on T2I-CompBench++~\citep{huang2025t2i}, covering attribute binding, spatial relations, and complex compositions; (2)~object-level accuracy on GenEval~\citep{ghosh2023geneval}, covering object presence, counting, and color; and (3)~dense prompt following on DPG-Bench~\citep{hu2024ella}.

\section{Experiments}

\subsection{CAPEval Metrics Better Explain Downstream Utility than Scale}
\label{sec:downstream}

\begin{figure}[t]
\centering
\includegraphics[width=1\linewidth]{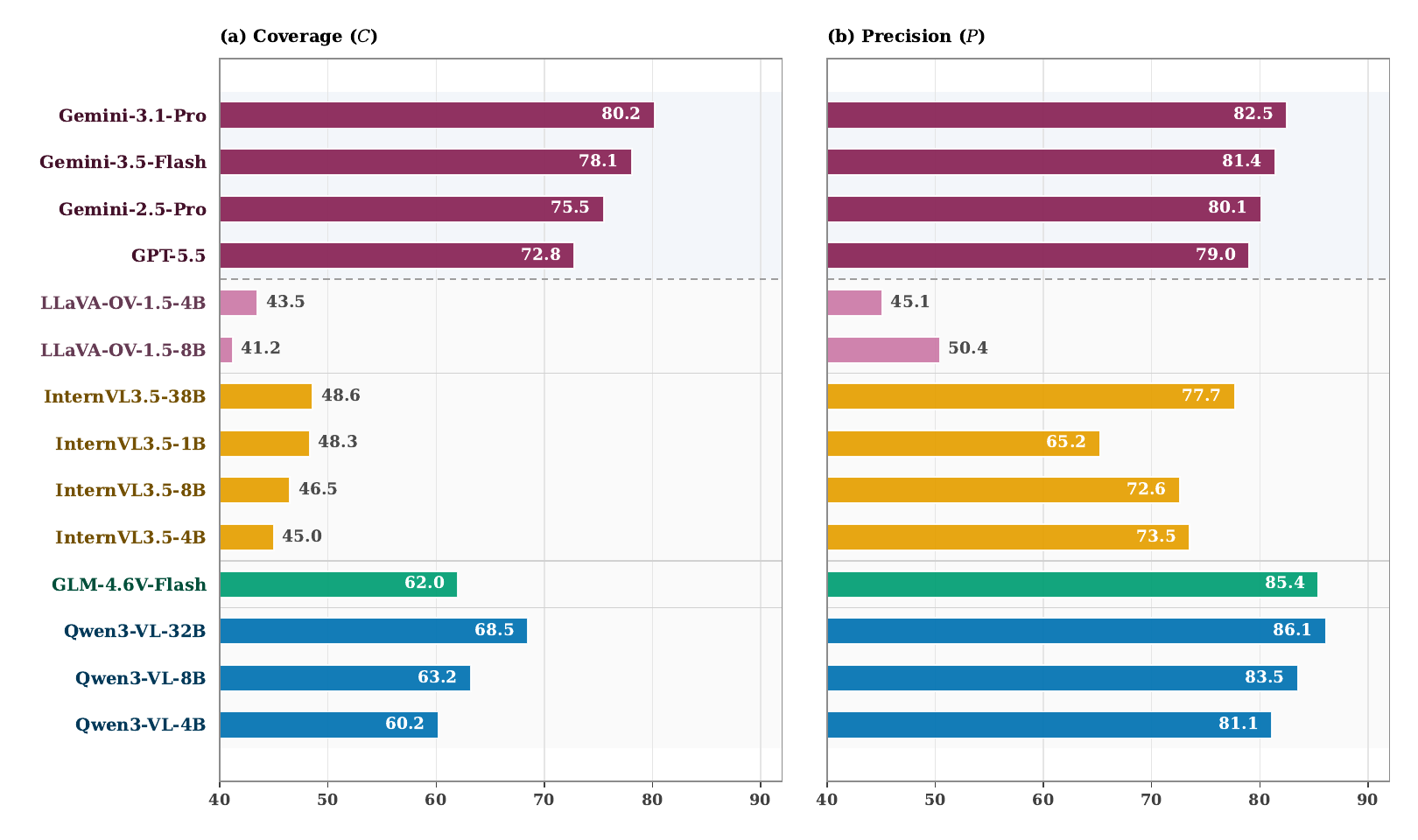}
\vspace{-20pt}
\caption{
Coverage ($C$) and Precision ($P$) scores of multimodal models on CAPEval.
The dashed horizontal line separates proprietary models from open-source models.
Captioners exhibit diverse and non-monotonically aligned Coverage--Precision profiles, showing that the two dimensions capture distinct properties of caption quality.
}
\label{fig:capeval_scores}
\vspace{-10pt}
\end{figure}

\textbf{CAPEval Reveals Decoupled Coverage and Precision Profiles.}
Figure~\ref{fig:capeval_scores} reports the Coverage and Precision scores of different captioners on CAPEval and reveals substantial variation along both dimensions.
The proprietary models~\citep{comanici2025gemini, gemini31pro, gemini35flash, gpt55} generally attain high scores on both dimensions.
Among the evaluated open-source captioners, Coverage ranges from $41.2$ to $68.5$, while Precision ranges from $45.1$ to $86.1$, indicating that captioners differ both in the breadth of visual information they describe and in the reliability with which that information is grounded in the image.

Importantly, Coverage and Precision exhibit \textit{no} monotonic relationship across captioners.
While some captioners achieve high scores on both dimensions, others show divergent quality profiles: some provide broader descriptions with lower factual reliability, \textit{e.g.}, InternVL3.5-1B, whereas others produce more conservative captions with higher precision, \textit{e.g.}, InternVL3.5-4B and InternVL3.5-8B.
%
%
This divergence suggests that captioners \textit{differ} in how much visual information they attempt to capture and how reliably they ground that information, leading to distinct trade-offs between semantic breadth and factual accuracy.

\textbf{Downstream evaluation.}
To examine how these distinct quality profiles relate to downstream utility, we use the captions generated by each captioner as supervision for two understanding pipelines and two generation pipelines.
%
%
%
The \textit{Avg.} columns in Table~\ref{tab:all_downstream_scores} further average the two pipeline-level scores within each downstream objective.
Raw benchmark scores are provided in Appendix~\ref{app:Raw_Score}.

\begin{table}[t]
\centering\small
\caption{
CAPEval Coverage ($C$) and Precision ($P$) together with aggregate downstream scores.
Shaded rows correspond to the InternVL3.5 family members used for the within-family comparison. 
The results show that captioner scale alone does not determine downstream utility, while the preferred caption quality profile differs between understanding and generation.
Detailed performance for each understanding and generation benchmark is provided in Appendix~\ref{app:Raw_Score}.
}
\vspace{-8pt}
\label{tab:all_downstream_scores}

\setlength{\tabcolsep}{9pt}
\begin{tabular}{lcccccccc}
\toprule
\textbf{Captioner}
& \multicolumn{2}{c}{\textbf{CAPEval}}
& \multicolumn{3}{c}{\textbf{Understanding}}
& \multicolumn{3}{c}{\textbf{Generation}}
\\

\cmidrule(lr){2-3}
\cmidrule(lr){4-6}
\cmidrule(lr){7-9}

& $C$ & $P$
& SigLIP & CLIP & Avg.
& SD3.5 & QwenImg & Avg.
\\
\midrule

Qwen3-VL-32B
& 68.5 & 86.1
& 66.8 & 53.6 & 60.2
& 70.2 & 76.7 & 73.5
\\

Qwen3-VL-8B
& 63.2 & 83.5
& 66.6 & 55.1 & 60.9
& 69.7 & 76.4 & 73.1
\\

Qwen3-VL-4B
& 60.2 & 81.1
& 67.5 & 54.0 & 60.8
& 69.2 & 76.1 & 72.7
\\

\midrule

GLM-4.6V-Flash
& 62.0 & 85.4
& 66.8 & 55.1 & 61.0
& 69.3 & 75.5 & 72.4
\\

\midrule

InternVL3.5-38B
& 48.6 & 77.7
& 65.0 & 52.5 & 58.8
& 68.0 & 75.5 & 71.8
\\

\rowcolor{gray!15}
InternVL3.5-8B
& 46.5 & 72.6
& 65.2 & 49.3 & 57.3
& 66.5 & 75.4 & 71.0
\\

\rowcolor{gray!15}
InternVL3.5-4B
& 45.0 & \textbf{73.5}
& 65.2 & 49.7 & 57.5
& 67.7 & 75.4 & \textbf{71.6}
\\

\rowcolor{gray!15}
InternVL3.5-1B
& \textbf{48.3} & 65.2
& 64.2 & 52.7 & \textbf{58.5}
& 65.9 & 74.3 & 70.1
\\

\midrule

LLaVA-OV-8B
& 41.2 & 50.4
& 65.5 & 52.9 & 59.2
& 62.1 & 69.0 & 65.6
\\

LLaVA-OV-4B
& 43.5 & 45.1
& 66.0 & 50.6 & 58.3
& 61.9 & 68.9 & 65.4
\\

\bottomrule
\end{tabular}

\vspace{-10pt}
\end{table}

\textbf{A counterintuitive phenomenon within InternVL3.5.}
We first consider InternVL3.5-8B/4B/1B as a controlled comparison within a single model family.
A natural expectation is that increasing captioner scale would produce more useful supervision and therefore stronger downstream models.
However, Table~\ref{tab:all_downstream_scores} shows that downstream performance does not improve monotonically with captioner scale.
For understanding, the ranking is 1B $(58.5) >$ 4B $(57.5) >$ 8B $(57.3)$, whereas for generation, the ranking is 4B $(71.6) >$ 8B $(71.0) >$ 1B $(70.1)$.
Thus, the largest captioner may \textit{not} always be the most useful source of supervision for either downstream objective.

These downstream rankings exhibit different alignments with $C$ and $P$.
Among the three models, InternVL3.5-1B has the highest Coverage ($C=48.3$) and also achieves the strongest understanding performance.
InternVL3.5-4B, by contrast, has the highest Precision ($P=73.5$) and achieves the strongest generation performance.
We next explore whether this pattern extends beyond a single model family through systematic regression analysis across all evaluated captioners.

\subsection{Systematic Regression Analysis Across Captioners}

\textbf{Setup and interpretation.}
To examine whether the pattern observed within InternVL3.5 generalizes across captioners, we fit separate ordinary least squares (OLS) regressions for each of the four downstream pipelines.
For captioner $i$ and downstream pipeline $t$, we model
\begin{equation}
U_{i,t} \ \ \text{or}\ \ G_{i,t}
=
\beta_{0,t}
+
\beta_{C,t} C_i
+
\beta_{P,t} P_i
+
\epsilon_{i,t},
\end{equation}
where $C_i$ and $P_i$ denote the CAPEval Coverage and Precision scores of captioner $i$, respectively, and $U_{i,t}$ or $G_{i,t}$ denotes its downstream score for pipeline $t$.
Each regression therefore estimates the association between each CAPEval dimension and downstream performance.

\begin{figure}[t]
    \centering\small
    \includegraphics[width=\linewidth]{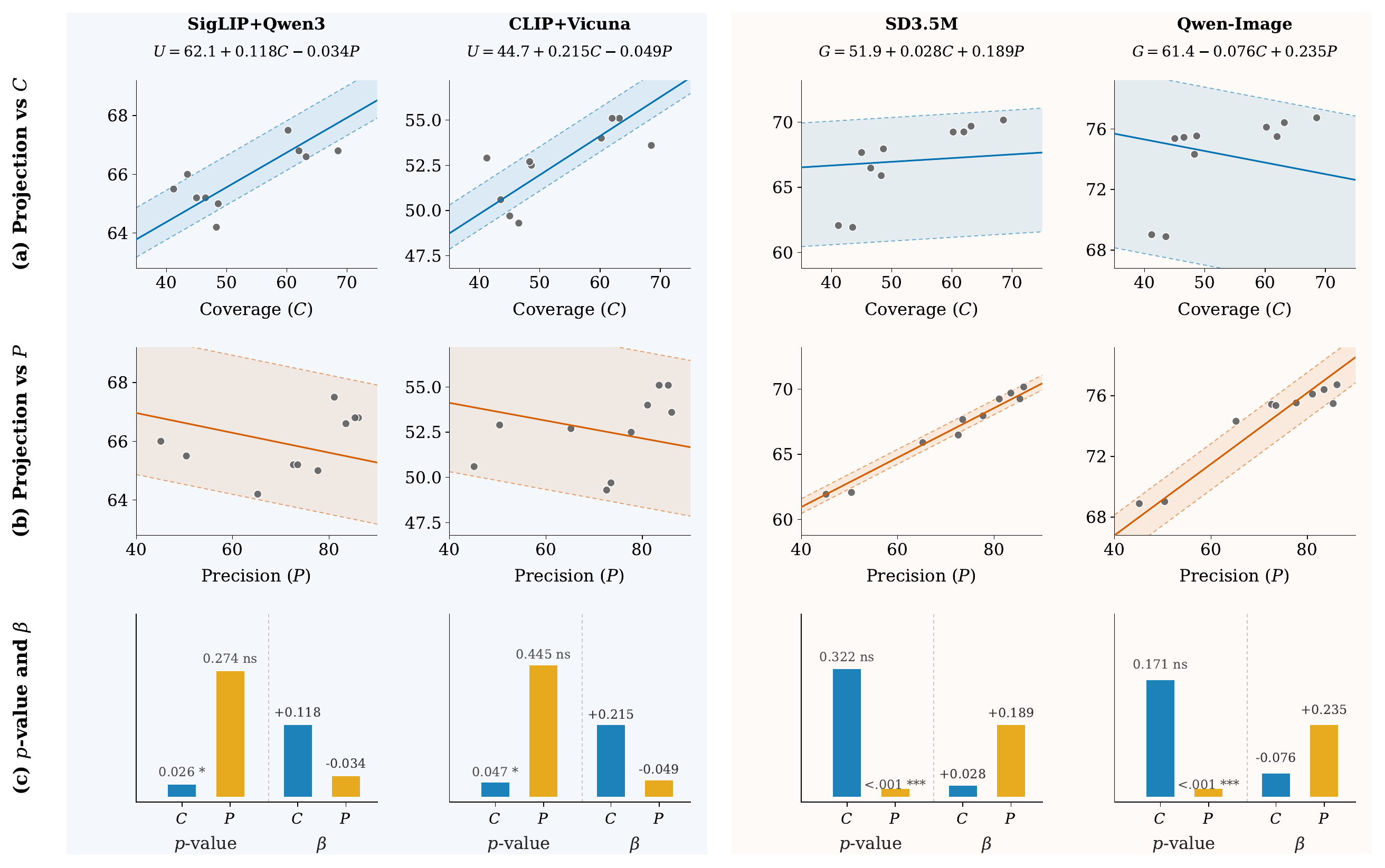}
    \vspace{-20pt}
    \caption{
    Regression analysis across four downstream pipelines.
    Row labels show the fitted equation for each pipeline.
    \textbf{(a)} Partial projection onto Coverage, with Precision fixed at its mean; the shaded region spans predictions obtained at the observed $P_{\min}$ and $P_{\max}$.
    \textbf{(b)} Partial projection onto Precision, with Coverage fixed at its mean; the shaded region analogously spans the observed $C_{\min}$ and $C_{\max}$.
    \textbf{(c)} OLS coefficient estimates and coefficient-level $p$-values for Coverage and Precision.
    For understanding, Coverage has a positive and significant coefficient in both SigLIP-Qwen3 and CLIP-Vicuna, whereas Precision is less significant.
    For generation, Precision is positive and significant in both SD3.5M and Qwen-Image, while Coverage is not.
    }
    \label{fig:regression}
    \vspace{-10pt}
\end{figure}

To jointly characterize the \textit{strength} and \textit{statistical reliability} of the link between caption quality and downstream performance, we center our analysis on two core statistics from the fitted models:
\begin{itemize}
    \item \textbf{The fitted coefficients \(\hat{\beta}_{C,t}\) and \(\hat{\beta}_{P,t}\)} denote the estimated change in downstream score corresponding to a one-unit increase in Coverage and Precision, respectively, for task $t$, which reflects the marginal contribution of each quality dimension to downstream utility.
    \item \textbf{The $p$-value for each coefficient} is computed under the null hypothesis \(H_0: \beta = 0\), and quantifies the statistical evidence that the observed conditional linear association differs from zero. 
    A smaller $p$-value indicates higher confidence that the correlation is statistically meaningful, rather than arising from random variation. 
\end{itemize}
Figure~\ref{fig:regression} visualizes the fitted models: (a) and (b) show partial projections onto Coverage and Precision, respectively, while (c) reports the estimated coefficients and coefficient-level $p$-values.

\begin{figure}
    \centering
    \includegraphics[width=1\linewidth]{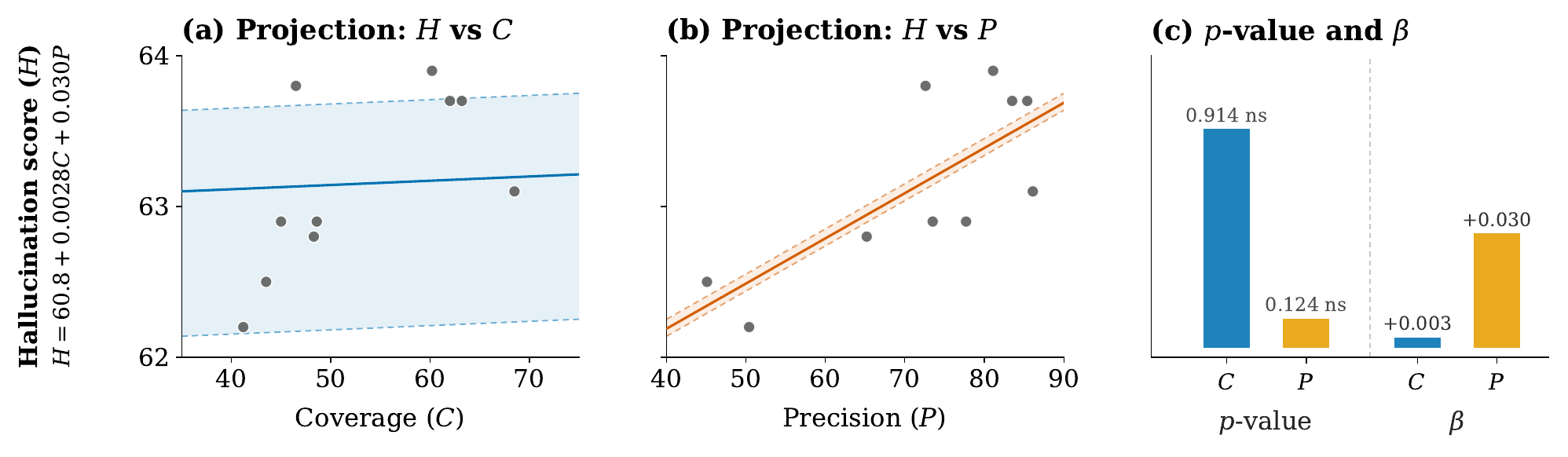}
    \vspace{-20pt}
    \caption{Regression analysis on \textit{hallucination} score shows the fitted equation for the average score of CLIP-Vicuna and SigLIP-Qwen3.
    For hallucination, Precision is more significant than Coverage.}
    \label{fig:hallucination}
\end{figure}

\textbf{Understanding favors broader semantic coverage.}
Across both understanding pipelines, Coverage is the statistically significant predictor of downstream performance, indicating that captions with broader semantic coverage provide more effective supervision for VLM understanding.
Specifically,
\begin{itemize}
    \item For SigLIP-Qwen3, $\hat{\beta}_C = +0.118$ with $p$-value $= 0.026$, while the effect of Precision is not statistically significant ($p$-value $= 0.274$).
    \item For CLIP-Vicuna, $\hat{\beta}_C = +0.215$ with $p$-value $= 0.047$, while the effect of Precision is not statistically significant ($p$-value $= 0.445$).
\end{itemize}
\textit{However, this pattern does not extend uniformly to hallucination performance.}
We further regress the average score of the two understanding pipelines across HallusionBench, POPE, and AMBER on Coverage ($C$) and Precision ($P$).
As shown in Figure~\ref{fig:hallucination}, Precision exhibits a stronger association with hallucination performance ($p$-value $= 0.124$) than Coverage ($p$-value $= 0.914$).
Thus, while general understanding primarily benefits from broader semantic Coverage, hallucination robustness is more closely associated with caption Precision.

\textbf{Generation favors factual precision.}
Across both generation pipelines, Precision is a positive and highly significant predictor of downstream performance, whereas Coverage provides no statistically significant additional explanatory power once Precision is controlled for.
Specifically,
\begin{itemize}
    \item For SD3.5M, $\hat{\beta}_P = +0.189$ with $p$-value $< 0.001$, while the effect of Coverage is not statistically significant ($p$-value $= 0.322$).
    \item For Qwen-Image, $\hat{\beta}_P = +0.235$ with $p$-value $< 0.001$, while the effect of Coverage is not statistically significant ($p$-value $= 0.171$).
\end{itemize}

\textbf{Caption utility is downstream-objective dependent.}
Taken together, the regressions reveal a clear asymmetry which is consistent with the InternVL3.5 case study.
For understanding, higher Coverage is associated with better performance, while Precision carries a weaker but negative association after controlling for Coverage.
For generation, the relationship reverses: Precision is the consistent predictor, whereas Coverage contributes little once caption reliability is accounted for.

This agreement between the within-family comparison and the cross-captioner analysis provides converging evidence that caption quality cannot be reduced to a single scalar notion.
Instead, different downstream objectives favor different caption quality profiles:
\textit{understanding benefits primarily from broader semantic Coverage, whereas generation depends more strongly on factual Precision.}
CAPEval makes this distinction explicit by decoupling these two properties, thereby providing a more informative characterization of caption utility.

\section{Conclusion}

In this paper, we introduced CAPEval, a framework that decomposes caption quality into two distinct axes: (1) Coverage, measuring how much of the visual content a caption attempts to describe, and (2) Precision, measuring how reliably those described facts are correct.
Moreover, CAPEval connects both axes to downstream multimodal performance through controlled training experiments. 
By keeping the image pool, model architecture, and training configuration fixed while varying only the caption source across ten captioners from four model families, CAPEval isolates the causal contribution of each caption quality dimension to vision-language understanding and text-to-image generation.
Our experiments reveal a consistent and task-dependent dissociation between the two axes.
For text-to-image generation, Precision is the only statistically significant predictor. 
For vision-language understanding, the pattern reverses: Coverage is the significant predictor, while Precision is not. 
This \textit{dissociation} surfaces a counterintuitive practical consequence: a smaller captioner can outperform a larger one when its Coverage--Precision profile better matches the target objective. 
Captioner scale alone is therefore an insufficient proxy for downstream utility.
We provide a task-aware view of caption quality and offer practical guidance for caption generation, filtering, and large-scale multimodal dataset construction: moving beyond single-score caption evaluation toward a principled, objective-driven framework.

{\small
\bibliographystyle{iclr2025_conference}
\bibliography{references}
}

\appendix

\section*{Appendix}

\section{Additional CAPEval Information}
\label{app:benchmark_stats}

\subsection{GT caption and Checklist of the Example}
\label{sec:gt_checklist_example}

\textbf{Ground-truth caption.}
In the foreground and mid-ground, various amusement park attractions create a vibrant scene. A purple roller coaster track with a ladder-like structure and two long black linear elements curves upward from the bottom center toward the right. Artificial palm trees with dark green fronds are distributed throughout the lower area; a large palm tree on the left partially obscures the view of the attractions behind it. To the far left, a low fence consists of vertical panels in alternating colors of blue, orange, teal, and magenta. Behind the palm tree on the far left, a yellow and blue ride structure with bucket-style seats is visible, resembling a small vertical Ferris wheel. In the center-left, a tall, intricate structure made of orange metal beams rises vertically, topped with a teal, pagoda-style roof with multiple internal staircases leading to a slide entrance on the right. Adjacent to it in the center is a teal framework supporting a platform where a group of people stands. Extending to the right, large tubular slides wind through the space; one section is orange, another is off-white, and another is green. In the lower right quadrant, a pool with a curved blue rim is filled with a white, snow-like substance rather than water. Several inflatable inner tubes in colors including pink and blue rest on the white surface. A person wearing a long black coat, black pants, white shoes, and a light pink headscarf stands near the pool's edge in the bottom right, facing away from the camera. A small portion of another person wearing a winter hat and dark clothing is visible at the very bottom edge near the fence. Behind the slides and tracks in the mid-ground, a low wall is decorated with a colorful mural depicting a tropical beach scene with blue waves and human figures. Real palm trees stand in front of the wall, distinct from the painted mural. To the right, a ride vehicle shaped like a vintage carriage with a pink curved roof, decorative elements, and black-and-white wheels runs on tracks. In the background among bare tree branches stands a vertical tower structure with a white spherical top decorated with pink dots, a flat platform below, and a multi-colored vertical body below that. A building with pinkish walls and windows is visible on the far right. The background is dominated by a skyline of tall buildings. A large building with a grid-like facade of numerous rectangular windows and a stepped roofline stands prominently on the right side of the skyline. A tall, solid, rectangular skyscraper with a light grey or metallic surface rises next to it. A modern building with a blue glass curtain wall is situated behind the orange tower structure. A distinctive building with a geometric, angular shape and a diagonal cross-braced facade is visible in the distance to the left of the orange tower. A blocky, beige residential-style building with rows of windows stands on the far left of the skyline. The sky occupies the upper portion of the image, displaying a smooth gradient from pale blue at the top to a soft, warm pinkish-orange hue near the horizon on the right side, suggesting the lighting of dawn or dusk. The lighting is soft and diffuse, creating a calm atmosphere with low contrast.

\textbf{Instance checklist.}
\begin{enumerate}[label=\arabic*., series=checklist]
    \item \textbf{[object]} Is a purple roller coaster track mentioned in the caption?
    \item \textbf{[plant]} Does the caption mention the presence of artificial palm trees?
    \item \textbf{[object]} Is a low fence with multi-colored panels described?
    \item \textbf{[object]} Does the caption describe a ride structure with bucket-style seats?
    \item \textbf{[object]} Are large tubular slides mentioned in the description?
    \item \textbf{[object]} Does the caption mention a pool with a curved blue rim?
    \item \textbf{[object]} Are inflatable inner tubes mentioned in the caption?
    \item \textbf{[object]} Does the caption describe a mural depicting a tropical beach scene?
    \item \textbf{[object]} Is a ride vehicle shaped like a vintage carriage mentioned?
    \item \textbf{[building]} Does the caption describe a vertical tower structure in the background?
    \item \textbf{[building]} Are there mentions of a skyline of tall buildings in the background?
    \item \textbf{[characters]} Does the caption mention the presence of a person near a pool?
\end{enumerate}

\textbf{Attribute checklist.}
\begin{enumerate}[label=\arabic*., resume=checklist]
    \item \textbf{[color]} Is the roller coaster track described as being purple?
    \item \textbf{[shape]} Does the caption state that the roller coaster track is curved?
    \item \textbf{[color]} Are the fronds of the artificial palm trees described as dark green?
    \item \textbf{[color]} Does the caption specify that the fence panels have alternating colors of blue, orange, teal, and magenta?
    \item \textbf{[material]} Is the tall, intricate structure described as being made of orange metal beams?
    \item \textbf{[shape]} Is the roof on the tall structure described as being pagoda-style?
    \item \textbf{[color]} Does the caption mention that the tubular slides have orange, off-white, and green sections?
    \item \textbf{[texture]} Is the substance in the pool described as white and snow-like?
    \item \textbf{[color]} Are the inner tubes mentioned to be in colors including pink and blue?
    \item \textbf{[color]} Is the person's coat described as being long and black?
    \item \textbf{[color]} Is the person's headscarf described as light pink?
    \item \textbf{[shape]} Is the roof of the vintage carriage ride described as curved?
    \item \textbf{[shape]} Is the top of the background tower described as a white sphere?
    \item \textbf{[color]} Are pink dots mentioned as a decoration on the tower's top?
    \item \textbf{[texture]} Is one of the background buildings described as having a grid-like facade?
    \item \textbf{[material]} Is a modern building described as having a blue glass curtain wall?
    \item \textbf{[shape]} Is a distant building described as having a geometric, angular shape?
    \item \textbf{[color]} Does the caption mention the sky has a gradient from pale blue to pinkish-orange?
    \item \textbf{[orientation]} Is the person standing near the pool described as facing away from the camera?
\end{enumerate}

\textbf{Relation checklist.}
\begin{enumerate}[label=\arabic*., resume=checklist]
    \item \textbf{[spatial]} Does the caption state that the roller coaster track curves upward from the bottom center toward the right?
    \item \textbf{[occlusion]} Is a large palm tree on the left described as partially obscuring the view?
    \item \textbf{[spatial]} Is the yellow and blue ride described as being behind the palm tree on the far left?
    \item \textbf{[spatial]} Is the teal framework described as being adjacent to the orange tower in the center?
    \item \textbf{[composition]} Does the caption mention a group of people standing on a platform?
    \item \textbf{[spatial]} Are the inflatable inner tubes described as resting on the white surface within the pool?
    \item \textbf{[spatial]} Is a person described as standing near the pool's edge in the bottom right?
    \item \textbf{[spatial]} Are real palm trees mentioned to be standing in front of the wall with the mural?
    \item \textbf{[spatial]} Is the vertical tower structure located in the background among bare tree branches?
    \item \textbf{[spatial]} Is the modern building with a blue glass curtain wall situated behind the orange tower structure?
    \item \textbf{[spatial]} Does the caption state that the sky occupies the upper portion of the image?
\end{enumerate}

\textbf{Image checklist.}
\begin{enumerate}[label=\arabic*., resume=checklist]
    \item \textbf{[scene\_information]} Does the caption describe the foreground and mid-ground as a vibrant scene with amusement park attractions?
    \item \textbf{[scene\_information]} Is the background described as being dominated by a skyline of tall buildings?
    \item \textbf{[emotions\_and\_atmosphere]} Does the caption suggest a calm atmosphere?
    \item \textbf{[hue\_temperature]} Does the caption suggest the lighting is from dawn or dusk due to the warm pinkish-orange hue?
    \item \textbf{[light\_effect]} Is the lighting described as soft and diffuse?
    \item \textbf{[saturation\_contrast]} Does the caption mention that the image has low contrast?
    \item \textbf{[picture\_composition]} Does the caption detail the placement of elements across the foreground, mid-ground, and background?
\end{enumerate}

\textbf{Human checklist.}
\begin{enumerate}[label=\arabic*., resume=checklist]
    \item \textbf{[human\_dressing]} Does the caption mention a person wearing a long black coat and black pants?
    \item \textbf{[human\_dressing]} Is a person described as wearing a light pink headscarf?
    \item \textbf{[human\_action\_position]} Is a person described as standing near the pool's edge?
    \item \textbf{[human\_action\_position]} Does the caption state that the person near the pool is facing away from the camera?
    \item \textbf{[human\_dressing]} Does the caption mention another person wearing a winter hat?
    \item \textbf{[human\_role]} Does the caption mention a group of people standing on a platform?
    \item \textbf{[human\_role]} Are human figures mentioned as being part of the colorful mural?
\end{enumerate}

\textbf{Text checklist.}
No text checklist items are provided for this example.

\textbf{UI checklist.}
No UI checklist items are provided for this example.

\textbf{World knowledge checklist.}
No world knowledge checklist items are provided for this example.

\subsection{Image Resolution Distribution}
\label{app:Image Resolution Distribution}

\begin{figure}[t]
\centering
\includegraphics[width=0.8\linewidth]{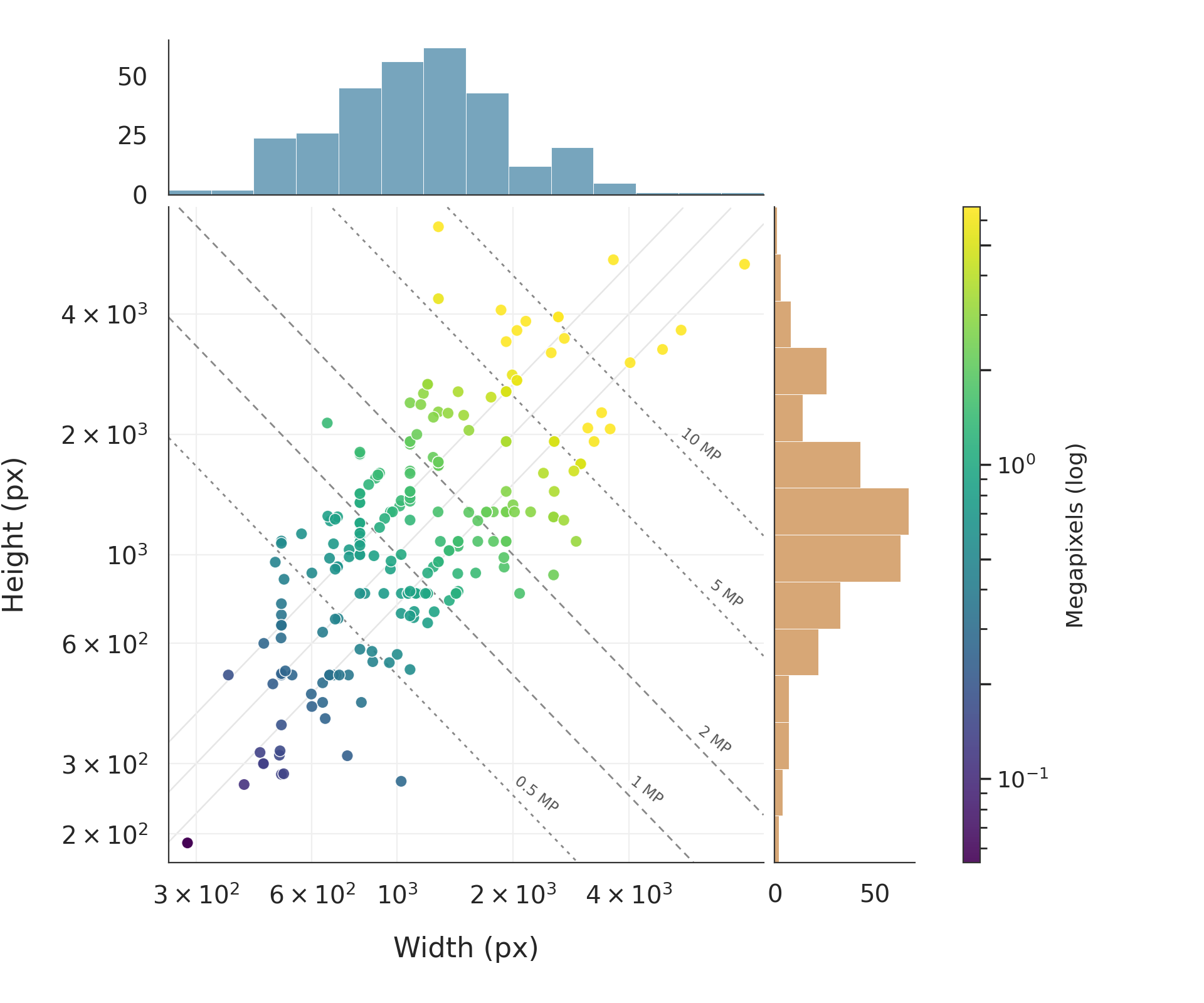}
\vspace{-15pt}
\caption{Distribution of image resolutions (height $\times$ width in pixels) across the 300 images in CAPEval. The benchmark includes images of various aspect ratios to ensure broad coverage of visual content types.}
\label{fig:resolution_distribution}
\end{figure}

Figure~\ref{fig:resolution_distribution} shows the distribution of image resolutions in the CAPEval benchmark. The images span a wide range of aspect ratios and resolutions, reflecting the diversity of real-world visual content.

\subsection{Checklist dimension}
\label{app:Checklist dimension}

Table~\ref{tab:dimension_statistics} summarizes the statistics of CAPEval's checklist annotations. Overall, CAPEval contains 14,965 verified atomic facts across 300 images, with an average of 49.88 checklist items per image. Instance, attribute, relation, and image-level dimensions are covered across all images, while other dimensions capture more specialized visual information when applicable. This diverse and dense annotation structure enables comprehensive evaluation of caption quality beyond object recognition, including fine-grained attributes, interactions, textual content, and contextual knowledge.

\begin{table}[t]
\centering\small
\caption{
Per-dimension statistics of checklist annotations in CAPEval, including the total number of verified atomic items, average number of items per image, and image coverage for each semantic dimension.
}
\vspace{-5pt}
\label{tab:dimension_statistics}
\begin{tabular}{lccc}
\toprule
\textbf{Semantic Dimension} & \textbf{Total Count} & \textbf{Avg. per Image} & \textbf{Images Covered} \\
\midrule
Instance        & 2,681 & 8.94  & 300 (100.0\%) \\
Attribute       & 3,660 & 12.20 & 300 (100.0\%) \\
Relation        & 2,632 & 8.77  & 300 (100.0\%) \\
Image           & 2,074 & 6.91  & 300 (100.0\%) \\
Text            & 1,800 & 6.00  & 217 (72.3\%) \\
Human           & 783   & 2.61  & 165 (55.0\%) \\
UI              & 584   & 1.95  & 123 (41.0\%) \\
World Knowledge & 751   & 2.50  & 218 (72.7\%) \\
\midrule
\textbf{Total} & \textbf{14,965} & \textbf{49.88} & \textbf{300 (100.0\%)} \\
\bottomrule
\end{tabular}
\vspace{-10pt}
\end{table}

\subsection{GT Caption and Checklist Annotation}
\label{app:annotation}

\textbf{Ground-Truth Caption Annotation.}
Each image is annotated by a trained human annotator who writes a comprehensive
English caption as a single unbroken paragraph. Annotators are instructed to
address the following dimensions \textit{only when the corresponding elements
are present in the image}:

\begin{itemize}
    \item \textit{Visual subjects.} Main subjects, quantity, position,
    interactions, and foreground/background scene context.
    \item \textit{Text content.} All legible text transcribed verbatim without
    translation, including layout, font, color, and text type (title, price,
    slogan, watermark, disclaimer).
    \item \textit{Aesthetic and photographic attributes.} Art style, color tone
    (saturation, contrast, temperature), lens type, shot size, light source,
    and shooting angle.
    \item \textit{Portrait attributes.} Apparent age bracket, gender, body
    shape, hairstyle, clothing and accessories, posture, and expression.
    Celebrity or public figure identities are noted when unambiguously
    identifiable.
    \item \textit{Special image properties.} Borders, watermarks, blur or
    distortion artifacts, composite or spliced images, and UI screenshot
    sources (app name noted where applicable).
\end{itemize}

The following constraints apply throughout:
\begin{itemize}
    \item Describe only objectively visible elements; do not speculate, infer emotional states, or introduce subjective evaluations.
    \item Do not fabricate details for elements that are not clearly visible in the image.
    \item State each fact exactly once; do not repeat descriptions of the same entity.\item Avoid vague or generic phrasing; use precise descriptions.
\end{itemize}

Each completed caption is reviewed by a second annotator for factual accuracy
and completeness; disagreements are resolved through discussion before
finalization.

\textbf{Checklist Annotation.}
Each GT caption is decomposed by human annotators into a structured checklist of atomic and verifiable statements. Every checklist item consists of two fixed fields:

\begin{itemize}
    \item \texttt{Tags}: a categorical label string identifying the specific concept or entity the item refers to.
    \item \texttt{Question}: a binary yes/no question verifiable directly against the image.
\end{itemize}

Items are organized across eight mandatory semantic dimensions:

\begin{itemize}
    \item \textit{Instance.} Object and entity identity, category, and count.
    \item \textit{Attribute.} Visual properties of entities: color, material, texture, shape, and size.
    \item \textit{Relation.} Spatial and functional relationships betweenentities.
    \item \textit{Image.} Scene-level properties: overall composition, image quality, lighting, and global visual style.
    \item \textit{Text.} All in-image text content, including signs, labels, watermarks, and interface strings.
    \item \textit{Human.} Person-related attributes: apparent demographics, clothing, pose, gesture, and expression.
    \item \textit{UI.} Interface elements such as buttons, menus, dialogs, icons, and layout components.
    \item \textit{World Knowledge.} Named entities, landmarks, brands, and external factual associations that require knowledge beyond visual inspection alone.
\end{itemize}

The schema is exhaustive and fixed: all eight dimensions are mandatory and no
custom dimensions may be added. Each item must be atomic (expressing exactly
one verifiable fact), grounded in both the GT caption and the image, and
assigned to exactly one dimension. Items requiring external knowledge are
assigned to \textit{World Knowledge}. A second annotator reviews all checklist
items for atomicity, factual grounding, and correct dimension assignment before
inclusion in the benchmark.

\begin{table}[t]
\centering\small
\caption{CAPEval scores of captioners. Scores are reported on a percentage scale. For each semantic dimension, C denotes Coverage, and P denotes Precision. Overall scores are highlighted in gray.}
\vspace{-10pt}
\label{tab:capeval_captioner_scores_cp_split}
\scriptsize
\setlength{\tabcolsep}{2.0pt}
\newcommand{\overallcell}[1]{\cellcolor{gray!15}#1}
\resizebox{\linewidth}{!}{
\begin{tabular}{ccccccccccc}
\toprule
\multirow{2}{*}{\textbf{Model}}
& \multicolumn{2}{c}{\overallcell{\textbf{Overall}}}
& \multicolumn{2}{c}{\textbf{Scene \& Object}}
& \multicolumn{2}{c}{\textbf{People \& Activity}}
& \multicolumn{2}{c}{\textbf{Text \& Interface}}
& \multicolumn{2}{c}{\textbf{Design \& Knowledge}} \\
\cmidrule(lr){2-3}
\cmidrule(lr){4-5}
\cmidrule(lr){6-7}
\cmidrule(lr){8-9}
\cmidrule(lr){10-11}
& \overallcell{C} & \overallcell{P}
& C & P
& C & P
& C & P
& C & P \\
\midrule
Gemini-3.1-Pro
& \overallcell{80.2} & \overallcell{82.5}
& 79.3 & 80.9
& 79.9 & 80.4
& 80.9 & 85.6
& 82.2 & 86.7 \\
Gemini-3.5-Flash
& \overallcell{78.1} & \overallcell{81.4}
& 77.5 & 79.2
& 76.1 & 81.4
& 78.8 & 82.9
& 81.4 & 85.8 \\
Gemini-2.5-Pro
& \overallcell{75.5} & \overallcell{80.1}
& 73.5 & 79.9
& 77.7 & 75.7
& 76.9 & 79.9
& 76.8 & 86.7 \\
GPT-5.5
& \overallcell{72.8} & \overallcell{79.0}
& 72.0 & 78.0
& 70.2 & 78.0
& 77.6 & 80.5
& 74.5 & 81.4 \\
\midrule
Qwen3-VL-32B
& \overallcell{68.5} & \overallcell{86.1}
& 70.3 & 84.6
& 66.8 & 85.9
& 66.2 & 88.0
& 68.1 & 89.3 \\
Qwen3-VL-8B
& \overallcell{63.2} & \overallcell{83.5}
& 63.8 & 82.3
& 65.0 & 80.5
& 60.3 & 88.6
& 61.7 & 86.6 \\
Qwen3-VL-4B
& \overallcell{60.2} & \overallcell{81.1}
& 61.5 & 80.9
& 58.1 & 79.9
& 58.8 & 83.9
& 60.7 & 80.6 \\
GLM-4.6V-Flash
& \overallcell{62.0} & \overallcell{85.4}
& 62.7 & 85.0
& 58.3 & 86.0
& 62.6 & 86.6
& 64.1 & 84.7 \\
InternVL3.5-38B
& \overallcell{48.6} & \overallcell{77.7}
& 48.9 & 75.9
& 51.2 & 76.6
& 46.2 & 81.3
& 46.6 & 81.4 \\
InternVL3.5-8B
& \overallcell{46.5} & \overallcell{72.6}
& 49.1 & 68.5
& 45.4 & 73.8
& 45.3 & 79.8
& 41.9 & 77.3 \\
InternVL3.5-4B
& \overallcell{45.0} & \overallcell{73.5}
& 45.8 & 70.4
& 44.5 & 72.7
& 45.2 & 79.6
& 43.4 & 78.0 \\
InternVL3.5-1B
& \overallcell{48.3} & \overallcell{65.2}
& 47.0 & 64.6
& 53.3 & 60.8
& 47.2 & 74.1
& 46.4 & 65.7 \\
LLaVA-OV-8B
& \overallcell{41.2} & \overallcell{50.4}
& 45.4 & 46.3
& 38.2 & 55.7
& 30.2 & 59.7
& 42.6 & 50.7 \\
LLaVA-OV-4B
& \overallcell{43.5} & \overallcell{45.1}
& 44.2 & 44.9
& 45.9 & 42.4
& 35.5 & 48.9
& 45.3 & 46.6 \\
\bottomrule
\end{tabular}
}
\vspace{-10pt}
\end{table}

\subsection{Scores of Downstream Understanding and Generation Benchmarks}
\label{app:Raw_Score}

To make heterogeneous benchmarks comparable, we map every raw benchmark score to a common $0$--$100$ scale before aggregation. The normalized understanding and generation benchmark scores are shown in Table~\ref{tab:full_understanding_generation_scores}.

\begin{table*}[t]
\centering
\vspace{-10pt}
\caption{
Normalized understanding and generation benchmark scores.
The CAPEval scores ($C$ and $P$), aggregate understanding score ($U$),
and aggregate generation score ($G$) are highlighted in gray.
}
\vspace{-6pt}
\label{tab:full_understanding_generation_scores}

\scriptsize
\setlength{\tabcolsep}{8pt}
\renewcommand{\arraystretch}{1.08}

\resizebox{\textwidth}{!}{%
\begin{tabular}{ccccccccccc}
\toprule

\multirow{2}{*}{\textbf{Metric}}
& \multicolumn{3}{c}{\textbf{Qwen3-VL}}
& \multicolumn{1}{c}{\textbf{GLM-4.6V}}
& \multicolumn{4}{c}{\textbf{InternVL3.5}}
& \multicolumn{2}{c}{\textbf{LLaVA-OV}}
\\

\cmidrule(lr){2-4}
\cmidrule(lr){5-5}
\cmidrule(lr){6-9}
\cmidrule(lr){10-11}

& \textbf{4B}
& \textbf{8B}
& \textbf{32B}
& \textbf{Flash}
& \textbf{1B}
& \textbf{4B}
& \textbf{8B}
& \textbf{38B}
& \textbf{4B}
& \textbf{8B}
\\


\midrule

\rowcolor{gray!15}
\textbf{$C$}
& \textbf{60.2}
& \textbf{63.2}
& \textbf{68.5}
& \textbf{62.0}
& \textbf{48.3}
& \textbf{45.0}
& \textbf{46.5}
& \textbf{48.6}
& \textbf{43.5}
& \textbf{41.2}
\\

\rowcolor{gray!15}
\textbf{$P$}
& \textbf{81.1}
& \textbf{83.5}
& \textbf{86.1}
& \textbf{85.4}
& \textbf{65.2}
& \textbf{73.5}
& \textbf{72.6}
& \textbf{77.7}
& \textbf{45.1}
& \textbf{50.4}
\\


\midrule
\multicolumn{11}{c}{\textbf{Understanding: CLIP - Vicuna}}
\\
\midrule

POPE
& 87.4 & 86.7 & 85.7 & 87.5 & 87.5 & 86.9 & 86.6 & 85.4 & 81.1 & 85.7
\\

Hall.
& 11.4 & 13.6 & 13.0 & 11.4 & 11.7 & 16.7 & 13.4 & 15.0 & 18.5 & 15.4
\\

AMBER
& 83.4 & 81.9 & 81.5 & 83.6 & 80.7 & 77.3 & 79.9 & 79.4 & 74.1 & 76.6
\\

MME
& 59.7 & 62.1 & 60.6 & 60.2 & 53.4 & 53.2 & 43.4 & 55.8 & 62.2 & 59.8
\\

MMB-EN
& 83.3 & 80.6 & 83.3 & 75.0 & 72.2 & 61.1 & 66.7 & 75.0 & 75.0 & 80.6
\\

MMB-CN
& 37.5 & 41.7 & 41.7 & 33.3 & 37.5 & 31.3 & 43.8 & 37.5 & 22.9 & 29.2
\\

ScienceQA
& 56.4 & 53.9 & 46.2 & 59.0 & 46.2 & 48.7 & 41.0 & 53.9 & 43.6 & 41.0
\\

MMStar
& 24.4 & 29.6 & 29.5 & 24.8 & 21.6 & 18.1 & 16.8 & 23.6 & 23.6 & 24.7
\\

OCR
& 34.6 & 34.4 & 34.7 & 36.5 & 37.2 & 35.5 & 34.5 & 35.0 & 34.5 & 35.0
\\

RWQA
& 42.8 & 47.5 & 41.2 & 53.9 & 49.7 & 43.1 & 33.5 & 46.8 & 42.4 & 50.6
\\

MMVP
& 57.0 & 60.3 & 53.7 & 63.0 & 60.0 & 45.7 & 37.7 & 50.7 & 54.3 & 61.0
\\

CV2D
& 59.8 & 62.8 & 61.5 & 60.4 & 60.0 & 55.5 & 61.8 & 59.9 & 57.0 & 63.1
\\

CV3D
& 60.4 & 62.7 & 59.3 & 66.8 & 59.5 & 59.0 & 64.3 & 56.3 & 59.2 & 60.5
\\

AI2D
& 56.6 & 56.3 & 57.1 & 56.4 & 56.0 & 56.6 & 55.4 & 55.2 & 54.2 & 56.4
\\

SEED
& 70.6 & 69.6 & 70.1 & 70.5 & 70.2 & 69.4 & 70.3 & 69.3 & 68.5 & 68.3
\\

MMMU
& 39.2 & 38.5 & 39.0 & 39.0 & 39.5 & 37.2 & 39.5 & 40.7 & 38.4 & 37.9
\\

\rowcolor{gray!15}
\textbf{$U$}
& \textbf{54.0} & \textbf{55.1} & \textbf{53.6} & \textbf{55.1} & \textbf{52.7}
& \textbf{49.7} & \textbf{49.3} & \textbf{52.5} & \textbf{50.6} & \textbf{52.9}
\\


\midrule
\multicolumn{11}{c}{\textbf{Understanding: SigLIP - Qwen3}}
\\
\midrule

POPE
& 87.6 & 88.6 & 88.5 & 87.7 & 87.5 & 87.9 & 87.2 & 85.1 & 88.9 & 87.7
\\

Hall.
& 28.1 & 30.8 & 24.6 & 25.7 & 27.0 & 28.1 & 31.4 & 27.5 & 27.7 & 27.7
\\

AMBER
& 85.4 & 80.5 & 85.0 & 86.2 & 82.6 & 80.8 & 84.3 & 84.8 & 84.7 & 80.3
\\

MME
& 68.9 & 68.2 & 66.8 & 67.7 & 67.4 & 69.0 & 66.2 & 66.2 & 68.5 & 69.6
\\

MMB-EN
& 88.9 & 94.4 & 91.7 & 94.4 & 91.7 & 86.1 & 83.3 & 88.9 & 88.9 & 91.7
\\

MMB-CN
& 85.4 & 81.3 & 79.2 & 75.0 & 68.8 & 68.8 & 79.2 & 68.8 & 83.3 & 77.1
\\

ScienceQA
& 66.7 & 69.2 & 66.7 & 69.2 & 56.4 & 56.4 & 48.7 & 51.3 & 64.1 & 62.1
\\

MMStar
& 42.9 & 41.2 & 42.5 & 40.4 & 42.4 & 42.7 & 43.6 & 46.4 & 43.2 & 41.8
\\

OCR
& 45.6 & 45.0 & 44.3 & 45.4 & 44.1 & 45.1 & 43.5 & 45.1 & 44.4 & 44.4
\\

RWQA
& 62.0 & 64.4 & 64.8 & 64.8 & 61.7 & 63.1 & 62.8 & 63.4 & 61.7 & 61.2
\\

MMVP
& 75.8 & 77.3 & 78.7 & 79.7 & 74.6 & 75.3 & 75.7 & 74.3 & 74.0 & 74.3
\\

CV2D
& 71.8 & 71.0 & 71.3 & 67.5 & 67.7 & 70.5 & 69.0 & 69.1 & 67.9 & 69.3
\\

CV3D
& 75.4 & 72.7 & 71.1 & 73.0 & 67.9 & 76.2 & 76.6 & 76.2 & 68.1 & 69.1
\\

AI2D
& 71.9 & 65.0 & 70.8 & 70.9 & 68.8 & 70.3 & 70.2 & 71.1 & 70.6 & 69.8
\\

SEED
& 75.3 & 69.6 & 75.5 & 76.0 & 74.2 & 74.4 & 75.0 & 75.0 & 73.9 & 74.0
\\

MMMU
& 48.5 & 46.1 & 47.9 & 44.9 & 44.9 & 47.7 & 46.7 & 46.8 & 46.5 & 46.8
\\

\rowcolor{gray!15}
\textbf{$U$}
& \textbf{67.5} & \textbf{66.6} & \textbf{66.8} & \textbf{66.8} & \textbf{64.2}
& \textbf{65.2} & \textbf{65.2} & \textbf{65.0} & \textbf{66.0} & \textbf{65.5}
\\


\midrule
\multicolumn{11}{c}{\textbf{Generation: SD3.5M}}
\\
\midrule

GenEval
& 70.5 & 70.9 & 71.6 & 70.4 & 64.4 & 68.1 & 65.9 & 67.5 & 59.1 & 58.4
\\

DPG
& 85.0 & 85.1 & 85.0 & 85.0 & 83.7 & 83.7 & 83.3 & 84.6 & 80.8 & 81.6
\\

T2I
& 52.2 & 53.0 & 53.9 & 52.3 & 49.6 & 51.3 & 50.2 & 51.8 & 46.0 & 46.3
\\

\rowcolor{gray!15}
\textbf{$G$}
& \textbf{69.2}
& \textbf{69.7}
& \textbf{70.2}
& \textbf{69.3}
& \textbf{65.9}
& \textbf{67.7}
& \textbf{66.5}
& \textbf{68.0}
& \textbf{61.9}
& \textbf{62.1}
\\


\midrule
\multicolumn{11}{c}{\textbf{Generation: Qwen-Image}}
\\
\midrule

GenEval
& 87.0 & 87.4 & 88.7 & 86.7 & 85.0 & 87.1 & 86.0 & 86.7 & 76.0 & 77.2
\\

DPG
& 86.4 & 86.7 & 86.3 & 85.9 & 85.4 & 85.0 & 86.2 & 85.7 & 82.6 & 83.2
\\

T2I
& 54.9 & 55.1 & 55.2 & 53.9 & 52.6 & 54.1 & 54.1 & 54.2 & 48.1 & 46.7
\\

\rowcolor{gray!15}
\textbf{$G$}
& \textbf{76.1}
& \textbf{76.4}
& \textbf{76.7}
& \textbf{75.5}
& \textbf{74.3}
& \textbf{75.4}
& \textbf{75.4}
& \textbf{75.5}
& \textbf{68.9}
& \textbf{69.0}
\\

\bottomrule
\end{tabular}%
}

\end{table*}

\section{CAPEval Scores and Per-captioner Downstream Scores}
\label{app:scores}

\subsection{CAPEval scores}

Table~\ref{tab:capeval_captioner_scores_cp_split} reports Coverage and Precision
scores for all 14 captioners across four semantic domains.

\section{Training Configuration}
\label{app:training_config}

\subsection{Vision-Language Understanding}

Table~\ref{tab:vlm_training_config} details the training setup for both VLM pipelines. Both use 8 GPUs, DeepSpeed ZeRO-3, bf16 mixed precision, AdamW optimizer (lr$=2{\times}10^{-5}$, cosine schedule, warmup ratio 0.03), gradient checkpointing, and \texttt{model\_max\_length=2048}. The first stage (visual-language alignment) uses the fixed LLaVA 558K pretraining set for both pipelines. The second stage (supervised fine-tuning) combines the 665K instruction mixture with 1.24M captioner-specific captions, filtered by token length to ensure training stability.

\begin{table}[t]
\centering
\small
\setlength{\tabcolsep}{4pt}
\caption{VLM training configuration.}
\vspace{-10pt}
\label{tab:vlm_training_config}
\begin{tabular}{lllcccc}
\toprule
\textbf{Pipeline} & \textbf{Stage} & \textbf{Data} & \textbf{Global BS} & \textbf{Per-dev} & \textbf{Accum} & \textbf{Steps} \\
\midrule
\multirow{2}{*}{CLIP--Vicuna}
  & 1 (align) & LLaVA 558K           & 256 & 8 & 4 & ${\sim}2{,}180$ \\
  & 2 (SFT)   & 665K+1.24M & 128 & 4 & 4 & ${\sim}14{,}900$ \\
\midrule
\multirow{2}{*}{SigLIP--Qwen3}
  & 1 (align) & LLaVA 558K           & 256 & 8 & 4 & ${\sim}2{,}180$ \\
  & 2 (SFT)   & 665K+1.24M & 64  & 2 & 4 & ${\sim}29{,}800$ \\
\bottomrule
\end{tabular}
\end{table}

\textbf{Vision encoders and LLMs:} CLIP uses ViT-L/336px with Vicuna-7B; SigLIP uses SigLIP-SO400M/384px with Qwen3-4B. \textbf{Projector:} CLIP warm-starts from the pretrained LLaVA MLP adapter; SigLIP trains the projector from scratch. \textbf{ViT tuning:} layers $\ge 12$ are unfrozen during alignment.

\subsection{Text-to-Image Generation}

Table~\ref{tab:t2i_training_config} details the training setup for both T2I suites. Both use DeepSpeed (ZeRO-2 for SD3.5M, ZeRO-3 with optimizer offload for QwenImage), bf16 mixed precision, AdamW ($\beta_1{=}0.9$, $\beta_2{=}0.999$, weight decay $10^{-4}$), cosine LR schedule, gradient clipping (max norm 1.0), resolution 1024, max sequence length 512, and global batch size 64. Only the diffusion transformer (DiT) parameters are trainable; text encoders and VAE are frozen, with embeddings or latents precomputed.

\begin{table}[t]
\centering
\small
\caption{T2I training configuration.}
\vspace{-10pt}
\label{tab:t2i_training_config}
\begin{tabular}{lcccccc}
\toprule
\textbf{Suite} & \textbf{Images} & \textbf{LR} & \textbf{Steps} & \textbf{Warmup} & \textbf{Loss Weight} & \textbf{DeepSpeed} \\
\midrule
SD3.5M      & 50K  & $5{\times}10^{-6}$ & 1600 & 100 & logit-normal & ZeRO-2 \\
QwenImage   & 100K & $1{\times}10^{-5}$ & 3200 & 200 & none & ZeRO-3 \\
\bottomrule
\end{tabular}
\vspace{-10pt}
\end{table}

\textbf{Preprocessing:} SD3.5M precomputes T5 and CLIP text embeddings; QwenImage precomputes both VAE latents and text embeddings.

\end{document}